\documentclass[10pt,twocolumn,letterpaper,table]{article}

\usepackage{wacv}
\usepackage{times}
\usepackage{epsfig}
\usepackage{graphicx}
\usepackage{caption}
\usepackage{subcaption}
\usepackage{amsmath}
\usepackage{amssymb}
\usepackage{tabularx}
\usepackage{booktabs}
\usepackage{bbding}
\usepackage{xcolor}
\usepackage{graphics}
\usepackage[accsupp]{axessibility}

\def\wacvPaperID{717} 

\wacvapplicationstrack 

\wacvfinalcopy 

\ifwacvfinal
\usepackage[breaklinks=true,bookmarks=false]{hyperref}
\else
\usepackage[pagebackref=true,breaklinks=true,colorlinks,bookmarks=false]{hyperref}
\fi

\begin{document}

\title{BirdSAT: Cross-View Contrastive Masked Autoencoders for Bird Species Classification and Mapping}

\author{Srikumar Sastry, Subash Khanal, Aayush Dhakal, Di Huang, Nathan Jacobs\\
Washington University in St. Louis\\
\{{\tt\small s.sastry, k.subash, a.dhakal, di.huang, jacobsn}\}{\tt\small @wustl.edu}
}

\maketitle

\begin{abstract}

We propose a metadata-aware self-supervised learning~(SSL)~framework useful for fine-grained classification and ecological mapping of bird species around the world. Our framework unifies two SSL strategies: Contrastive Learning~(CL) and Masked Image Modeling~(MIM), while also enriching the embedding space with metadata available with ground-level imagery of birds. We separately train uni-modal and cross-modal ViT on a novel cross-view global bird species dataset containing ground-level imagery, metadata (location, time), and corresponding satellite imagery. We demonstrate that our models learn fine-grained and geographically conditioned features of birds, by evaluating on two downstream tasks: fine-grained visual classification~(FGVC) and cross-modal retrieval. Pre-trained models learned using our framework achieve SotA performance on FGVC of iNAT-2021 birds and in transfer learning settings for CUB-200-2011 and NABirds datasets. Moreover, the impressive cross-modal retrieval performance of our model enables the creation of species distribution maps across any geographic region. The dataset and source code will be released at~\url{https://github.com/mvrl/BirdSAT}.
\end{abstract}

\section{Introduction}
Species classification and distribution mapping are two important tasks for ecologists who monitor and protect the habitats of endangered species. Species classification involves categorizing species with subtle differences into fine-grained classes. It lies within the continuous manifold of basic visual recognition tasks and more complex visual identification tasks. Moreover, it coincides with the task of Fine-Grained Visual Classification (FGVC) which has already been used for distinguishing between models of cars~\cite{Akrause20133d}, species of birds~\cite{van2018inaturalist,van2015building,wah2011caltech}, airplanes~\cite{maji2013fine}, etc. 
On the other hand, the task of species distribution mapping aims at mapping the habitation of species of interest over any geographic region in the world. In this work, we propose to learn a unified representation space useful for solving both of these tasks. Specifically, we evaluate our framework for classifying and mapping \texttt{bird} species around the world. However, our models are general enough to be easily extended to~\texttt{any} species of interest.

As easy as it may sound, low inter-class variance and high intra-class variance make the task of species classification relatively difficult. Most often, species in the same category vary in terms of their pose, size, and lighting. 
This makes it challenging for deep learning models to extract category-specific rich features useful for fine-grained classification and mapping.
Previous works have approached this challenge in one of the following ways: (1) Collecting additional labeled data~\cite{reed2016learning}; (2) Using sophisticated learning techniques~\cite{dubey2018pairwise,yang2018learning}; (3) Using auxiliary and/or metadata as additional cue~\cite{ACdiao2022metaformer,zhang2018fine}. Out of these, (1) is usually the most time-consuming and expensive approach and (2) requires careful design of objectives and methods for effective results. However, the inclusion of metadata has proven to be very effective. 

The task of species classification requires fine-grained visual representation learning capabilities. Recently, self-supervised learning (SSL) strategies such as Masked Image Modeling~(MIM)~\cite{he2022masked,satmae2022} have proven to be useful for learning discriminative features. On the other hand, the task of species mapping, which can be realized as a cross-modal retrieval task, would benefit from contrastive learning (CL) based SSL. To learn a common embedding space for both of these tasks, we propose to use a general SSL framework trained using objective functions for both MIM and CL. 

\begin{figure*}[!t]
\begin{center}
   \includegraphics[width=0.9\linewidth]{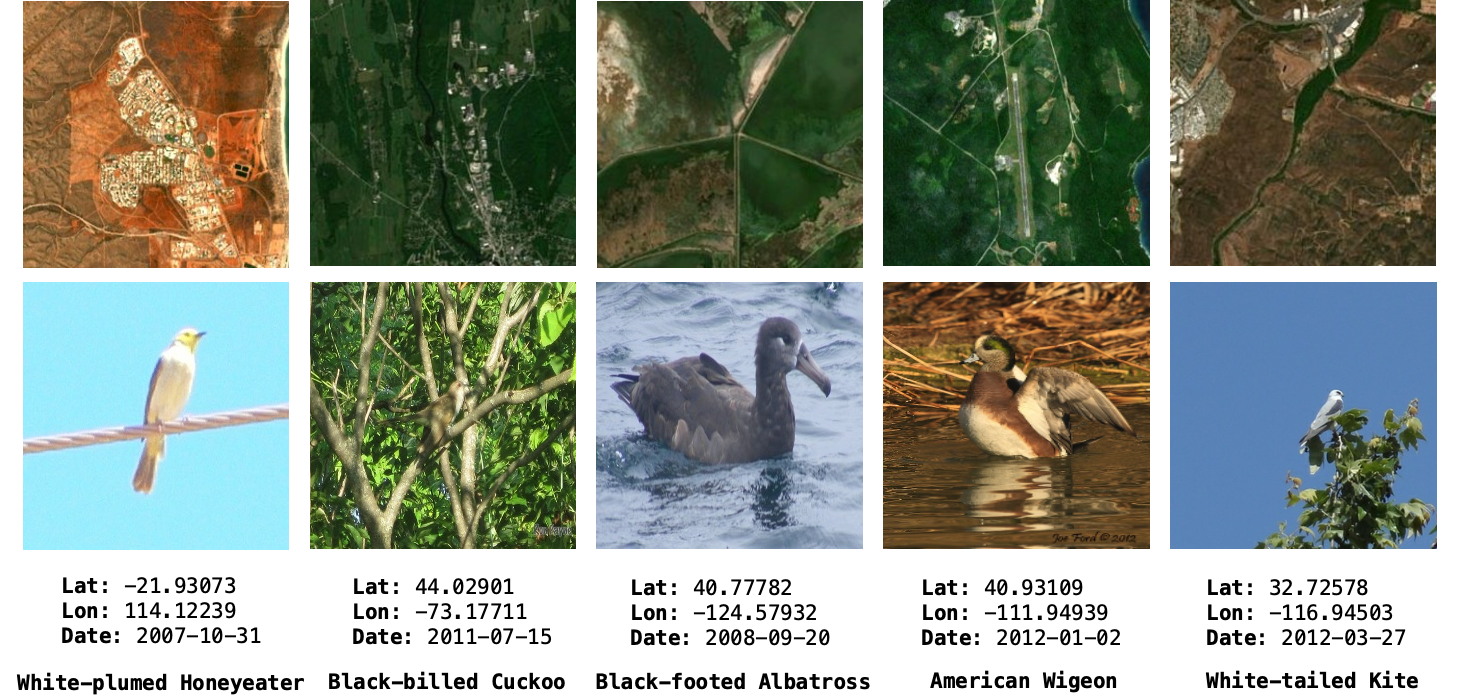}
\end{center}
   \caption{\textbf{Cross-View iNAT-2021 Birds Dataset}. Examples of paired satellite and ground-level images of birds along with metadata associated with each pair.}
\label{fig:data}
\end{figure*} 

We expect geolocation and time to provide useful cues for species classification and mapping. Therefore, we incorporate metadata (location, time) into our SSL framework, as additional information to learn from. However, metadata alone is not sufficient for species mapping as noted by other works on geo-aware mapping tasks~\cite{khanal2023learning,dhakal2023sat2cap}. Therefore, we additionally incorporate cross-view visual information by collecting freely available corresponding satellite images for each ground-level image. We expect that these images shall provide the model with a context of the surroundings and habitat a bird might be found. Further, information coming from multiple modalities is usually combined using early-fusion or late-fusion style architectures. Early-fusion style architectures~\cite{bachmann2022multimae,tang2022tvlt} use a multi-modal model to encode all input information while the latter uses modality-specific models to learn correlated information between the modalities~\cite{he2020momentum,radford2021learning}. In this work, we explore and evaluate these kinds of architectures from heuristic and systematic perspectives. 
In the end, using the models, we are able to map species of birds across the globe at a fine-grained level.



\begin{figure*}[!t]
\centering
\begin{subfigure}{.47\linewidth}
  \centering
  \includegraphics[width=\columnwidth]{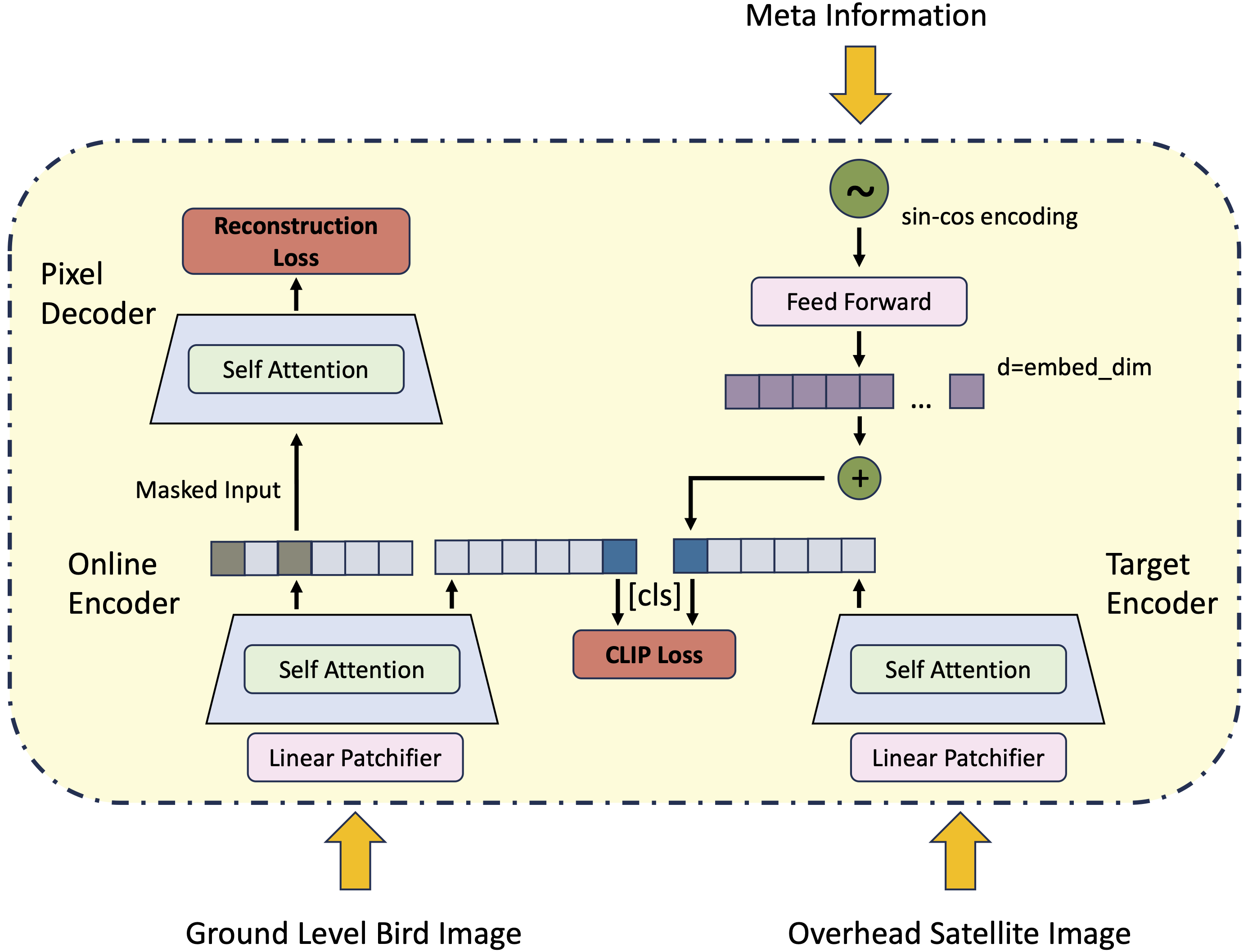}
  \caption{Cross-View Embed MAE}
  \label{fig:sub1}
\end{subfigure}%
\begin{subfigure}{.53\linewidth}
  \centering
  \includegraphics[width=\columnwidth]{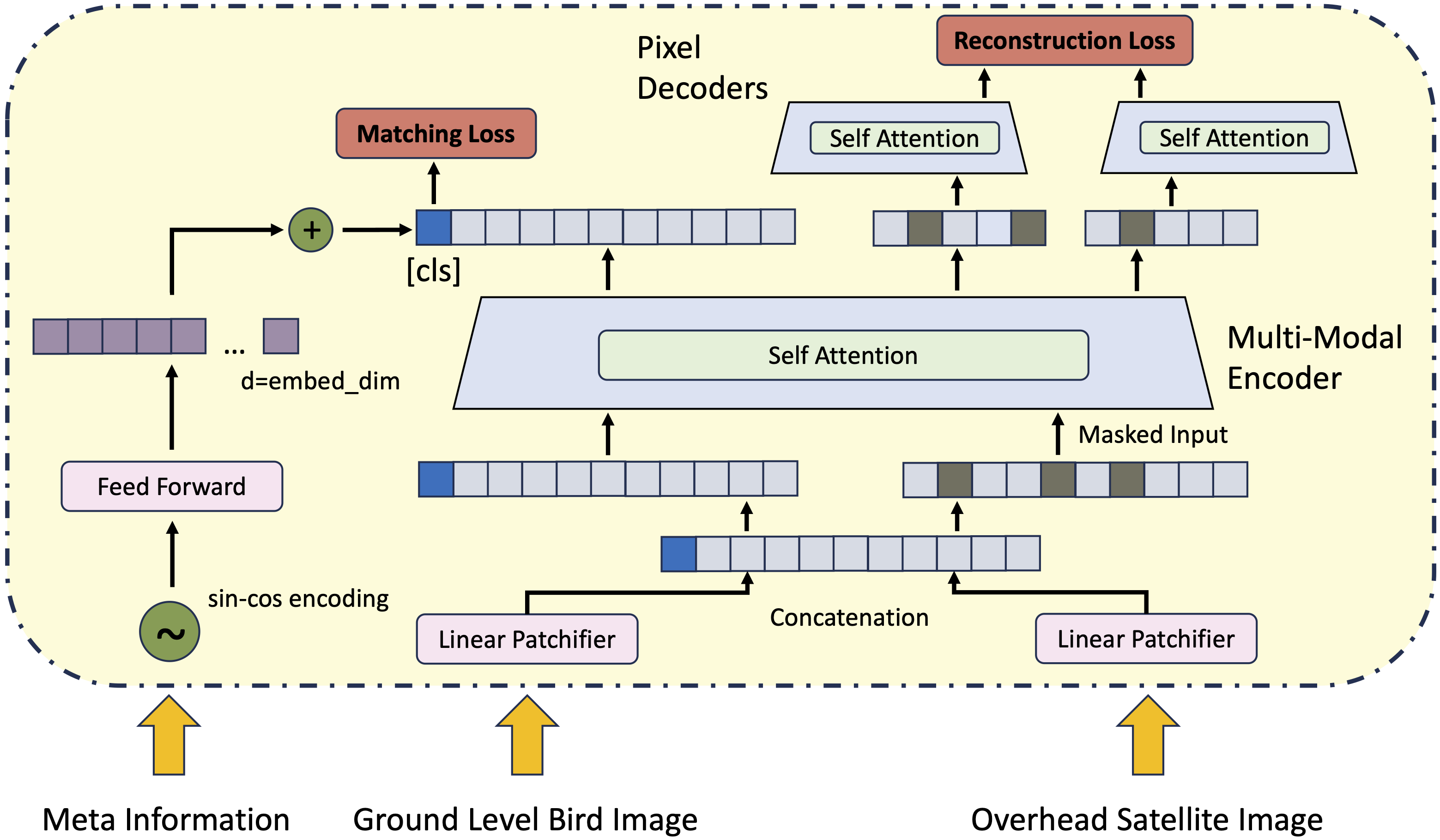}
  \caption{Cross-View Metric MAE}
  \label{fig:sub2}
\end{subfigure}
\caption{\textbf{Our proposed framework}. We evaluate (a) uni-modal (late-fusion) and (b) cross-modal (early-fusion) pre-training of ViT incorporating metadata and contrastive and masked reconstruction objectives.}
\label{fig:test}
\end{figure*}
\noindent
The contributions of our work are threefold:

\begin{itemize}
    \item We introduce a global Cross-View iNAT 2021 Birds Dataset, which contains paired satellite images and corresponding ground-level bird images.
    \item We propose a framework for cross-view pre-training of vision transformers along with metadata enabling the ecological mapping of bird species.
    \item We demonstrate the rich representational capability of our pre-trained models by demonstrating SotA on fine-grained bird classification across three datasets.
\end{itemize}

\section{Related Work}
\textbf{Self-Supervised Learning (SSL)} has proven to be an effective pre-training strategy for various downstream tasks in computer vision. The two most successful SSL strategies are: Contrastive Learning (CL) and Masked Image Modeling (MIM). Contrastive learning-based pre-training pulls positive pair of samples closer while pushing the negative pairs farther in the embedding space. CL creates a representation space with high instance discriminability useful for various visual recognition tasks. On the other hand, inspired by Masked Language Modeling (MLM)~\cite{devlin2018bert} in Natural Language Processing (NLP), MIM has proven to be an effective pre-training strategy in computer vision. MIM is effective, especially for tasks where learning fine-grained concepts is important. MIM was first introduced for vision tasks by Masked Autoencoder (MAE)~\cite{he2022masked}. MAE has since been used as a representation learning framework for video~\cite{tong2022videomae}, as well as for other visual modalities such as satellite imagery~\cite{satmae2022,reed2022scale}. Moreover, owing to its flexibility and scalability, MAE has also been adapted for different multi-modal representation learning frameworks such as MultiMAE~\cite{bachmann2022multimae} and M3AE~\cite{geng2022multimodal}. MAE frameworks offer rich representational capability useful for fine-grained tasks, however, the limited discriminability of its embedding space hampers the performance on visual recognition tasks. To mitigate this, some of the recent works~\cite{cmae:huang2022contrastive,cmaev:lu2023cmae,cmid:muhtar2023cmid} have proposed to introduce CL-style learning into the MIM-based MAE framework. One of the main downstream tasks of our work is fine-grained visual recognition. Therefore, we also pre-train our proposed cross-view framework using both contrastive and MIM losses.

\textbf{Fine-Grained Visual Classification~(FGVC)} requires distinguishing subtle yet discriminative details within a category (e.g., animal, bird, car, etc.). Accordingly, most of the prior works have proposed different attention mechanisms~\cite{li2017dynamic,fu2017look, zheng2017learning,zheng2019looking} which detect the discriminative parts in an image and amplify their corresponding features for recognition. Moreover, in order to enhance fine-grained representations, different modules that are easy to be plugged into common backbone architectures have been proposed~\cite{yang2018learning,chou2023fine}. In a separate line of work, different SSL techniques have been introduced either as an additional self-supervision~\cite{yu2022spare} or as a pre-training strategy for FGVC. For example, Yu et.al.~\cite{yu2022spare} propose randomly masking parts of an image and forcing the network to predict the position of the masked parts. Different pre-text tasks for SSL such as jigsaw solving, adversarial learning, and SimCLR~\cite{chen2020simple} based CL are explored in~\cite{breiki2021self}. In~\cite{yang2021self}, a multi-stage SSL strategy is proposed, where a SimCLR-style framework is trained with images progressively degraded with masks having different granularity at each stage. In a recent work~\cite{shu2023learning} an additional GradCAM-guided loss is introduced into a MoCo-style SSL framework. In our work, inspired by the success of SSL in various computer vision tasks including FGVC, we propose a cross-view SSL framework trained on our novel dataset containing ground-level images paired with their corresponding satellite imagery.

\textbf{Geography-Aware Learning} leverages the high-level context available in the geolocation of any ground-level scene. Such information proves to be a valuable signal for various visual recognition tasks~\cite{tang2015improving,mai2023csp,chu2019geo,ayush2021geography} and has been successfully used for mapping the distribution of different attributes across a geographic region of   interest~\cite{greenwell2018goes,salem2020learning,khanal2023learning,dhakal2023sat2cap}. For example, Tang et.al.~\cite{tang2015improving} proposed to encode location information into their network yielding improved visual recognition performance. Ayush et.al.~\cite{ABayush2021geography} proposed adding a geolocation classification loss into the original MoCo-v2 SSL framework achieving performance gain in a diverse range of remote sensing tasks. Similarly, from some of the recent works~\cite{chu2019geo,mai2023csp}, it has become evident that including geographic information improves the performance on the task of FGVC. Inspired by these findings, in both of the cross-view SSL frameworks proposed in our work, we encode location and date as extra metadata that the model can learn from.
\section{Cross-View iNAT-2021 Birds Dataset}
We construct a cross-view birds dataset that consists of paired ground-level bird images and satellite images as shown in Figure~\ref{fig:data}. We expect that this kind of dataset will not only help improve the performance of existing methods but also enable innovative new methods for bird distribution modeling. To do this, we select the iNAT-2021 dataset~\cite{van2018inaturalist} which spans all over the globe. This dataset is both large scale and contains rich metadata such as geolocation and timestamp of an image.

We carefully filter images of bird species from the dataset which have geolocation associated with them. This resulted in dropping only 888 out of 414,847 (0.2\%) observations in training. In testing, we dropped 29 out of 14,860 (0.1\%). This did not significantly impact the distribution of the classes (more details in Appendix Section A). Using the geolocation information, we collect Sentinel-2 level 2A images corresponding to each of the ground-level bird images. Each Sentinel-2 image we extract is of resolution 256x256 which spans an area of 6.55 km$^2$ on the Earth's surface. In total, the dataset contains 413,959 pairs for training and 14,831 pairs for testing.
\section{Method}
We employ and evaluate two different approaches for contrastively training MAE with satellite images and ground-level bird images. We describe the two approaches in the following sections.

\subsection{Cross-View Embed MAE}
The overall framework of Cross-View Embed MAE (CVE-MAE) is illustrated in Figure~\ref{fig:test} (a). Our method consists of two separate modality-specific transformer encoders and a single transformer decoder for reconstructing the ground-level image modality. Both the encoders have the same architecture based on ViTAE~\cite{xu2021vitae}, while the decoder has the same architecture as employed by the authors in MAE. The satellite image encoder is directly taken from~\cite{wang2022advancing} and is kept frozen throughout.

Similar to other contrastive learning frameworks~\cite{chen2020simple,he2020momentum}, the ground-level image encoder serves as the online encoder and the satellite image encoder serves as the target encoder. Both the encoders have an extra [cls] token which we use for computing a contrastive objective. The contrastive objective we use in this study is the symmetric InfoNCE loss as used in CLIP~\cite{radford2021learning}. If $I^g$ denotes ground-level image and $I^s$ denotes satellite image, CLIP loss is defined by:
\begin{equation}
    L_g = -log\frac{exp(I^g\cdot I^s)^+}{exp(I^g\cdot I^s)^++\sum_{j=1}^{N-1}exp(I^g\cdot I^s_j)^-}
\end{equation}
\begin{equation}
    L_s = -log\frac{exp(I^s\cdot I^g)^+}{exp(I^s\cdot I^g)^++\sum_{j=1}^{N-1}exp(I^s\cdot I^g_j)^-}
\end{equation}
\begin{equation}
    L_c = \frac{L_g+L_s}{2}
\end{equation}
Here, $I^g\cdot I^s$ is the normalized cosine similarity between the \textit{[cls]} token obtained from ground-level and satellite image encoders respectively. The sum is over a batch of samples and $+$ and $-$ denote positive and negative pairs within the batch respectively. Similar to MAE, the decoder is used to reconstruct ground-level images using masked versions of tokens obtained from the online encoder. A second forward pass is required to train for this objective as the ground-level encoder requires only the unmasked tokens as input. For the reconstruction objective, we use the $L_2$ loss defined by: 
\begin{equation}
    L_r = \sum_{j=1}^N|\hat{I^g_j} - I^g_j|^2_2
\end{equation}
The overall loss is then defined as -
\begin{equation}
    L = L_c + L_r
\end{equation}

Different from the existing method (i.e. CMAE~\cite{cmae:huang2022contrastive}), our method gets rid of the feature decoder layer. We found that training without the feature decoder layer results in stable loss curves.

\begin{table*}[!t]
\caption{Comparison of accuracy (\%) achieved by our proposed models and SotA approaches on the standard test set of the iNAT-2021 Birds dataset. We report linear probing (lin) and fine-tuning (ft) accuracy.}
\label{table-runtime}
\begin{center}
\begin{tabularx}{0.98\linewidth}{lccccccc}
Method & Location& Date & Pre-training & \#param. (trainable) &\#FLOPS &lin & ft \\
\midrule
MoCo-V2-Geo~\cite{ABayush2021geography} &\Checkmark&\XSolidBrush&InfoNCE+Geo-Clf.&115M&13.46G&52.44&85.07\\
MAE~\cite{he2022masked} &\XSolidBrush&\XSolidBrush&Recons. Loss&115M&13.46G&41.10&83.14\\
MetaFormer-2~\cite{ACdiao2022metaformer} & \Checkmark &\Checkmark&ImageNet Clf.&81M&16.90G&-&85.34 \\
\bottomrule
CVE-MAE & \XSolidBrush & \XSolidBrush&InfoNCE+Recons. Loss&117M&13.46G&38.86&83.78\\
CVE-MAE-Meta &\Checkmark &\Checkmark&InfoNCE+Recons. Loss&117M&13.46G&59.26&86.23\\
CVM-MAE &\XSolidBrush&\XSolidBrush&Matching+Recons. Loss&115M&31.59G&44.25&85.89\\
CVM-MAE-Meta &\Checkmark&\Checkmark&Matching+Recons. Loss&115M&31.59G&\cellcolor{gray!15}\textbf{63.33}&\cellcolor{gray!15}\textbf{87.46}\\
\end{tabularx}
\end{center}
\end{table*}

\subsection{Cross-View Metric MAE}
The overall framework of Cross-View Metric MAE (CVM-MAE) is illustrated in Figure~\ref{fig:test} (b). This kind of training strategy requires a single multi-modal transformer encoder and separate modality-specific transformer decoders. This is a similar setup as used by previous works on multi-modal MAE~\cite{geng2022multimodal,tang2022tvlt}. The encoder and decoders have the same architecture as employed by the authors in MAE.

The proposed framework starts by concatenating the tokens computed from the ground level and satellite images using separate linear patchifier layers. A [cls] token is appended to the tokens which is later used for computing the matching loss. \textbf{Ground-satellite Matching} predicts whether a pair of satellite and ground-level bird images is positive or negative. A single feed-forward layer is appended so as to train for the objective. The matching loss is simply defined as the binary cross entropy loss between ground-truth labels and the output of the feed-forward layer as follows:
\begin{equation}
L_m = \frac{-1}{2N}\sum_{j=1}^{2N}(y_jlog(\hat{y_j}) + (1-y_j)log(1-\hat{y_j}))
\end{equation}

Positive and negative pairs of ground-level and satellite images are defined using the batch currently in training. The satellite image batch is simply rolled to create the set of negative pairs. The output [cls] token from the multi-modal encoder is used for computing this matching loss. Intuitively, this token should capture the joint representation of the image pair.

Additionally, a second forward pass is required for computing the reconstruction objectives. This is necessary as the pixel decoders require a masked version of the encoder outputs. Unmasked tokens are first generated at the input stage using the patchifier layers. They are concatenated and sent to the multi-modal encoder. The tokens computed by the multi-modal encoder are separated back to their respective modality. Finally, the tokens (after concatenating with [mask] tokens) are sent to their modality-specific decoders for reconstruction. Again, we use the $L_2$ loss to train for the reconstruction objective. The overall loss is defined as - 
\begin{equation}
    L = L_m + L_r
\end{equation}
\subsection{Incorporating Acquisition Metadata}

While often ignored, acquisition metadata, such as when and where an image was captured, provides additional context which can improve our ability to interpret the content of an image. It can not only help reduce the number of possible classes but also help improve the interpretability of a model. Our Cross-View iNAT-2021 Birds Dataset provides geolocation and timestamp for each image. In our implementation, we use latitude, longitude, and month attributes of the metadata. As each of the attributes is a real number, we encode them using the sin-cos encoding method. They are then passed to a feed-forward layer which outputs an embedding of the same dimension as our single-stream and dual-stream encoders. Finally, they are added to the [cls] token embedding that results from the encoders. We call these models CVE-MAE-Meta and CVM-MAE-Meta. Note that for our dual stream approach, we only add metadata to the \textit{[cls]} token resulting from the satellite image encoder.

\begin{figure*}[!ht]
\begin{center}
   \includegraphics[width=\linewidth]{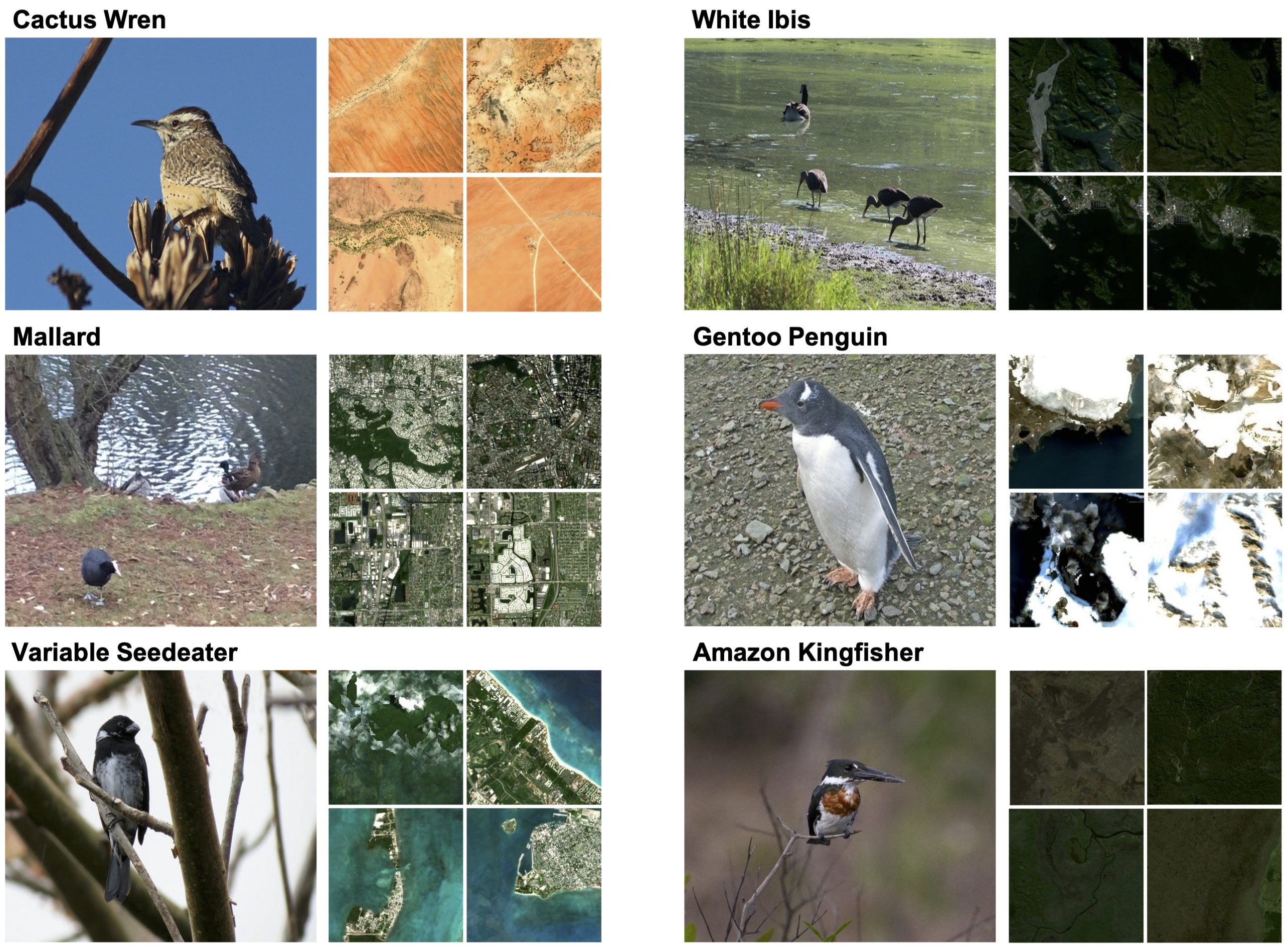}
\end{center}
   \caption{\textbf{Top-4 Retrieved Candidates}. We select six different bird species and retrieve the top-4 most similar satellite images (right) to the corresponding bird images (left) in the test set. Notice that the retrieved images are similar to each other while also being relevant for the corresponding bird species.}
\label{fig:birddist}
\end{figure*}

\subsection{Meta-Dropout}
During initial training runs for pre-training and fine-tuning, we noticed a heavy dependence of our models on metadata for minimizing target objectives (Appendix Section C). For several ground-level images of birds, our models seemed to ignore visual information completely. To address this issue, we randomly dropped metadata (25\% of the time) during training. Given the flexibility of our models, it is easy to forward the raw features without having the need to add metadata. Another benefit of this strategy is that it improves inference on unseen examples where metadata is not available.

\section{Experiments}
We evaluate the performance of our proposed models by first pre-training on the Cross-View iNAT-2021 Birds Dataset and then applying them to various tasks. In the following sections, we describe our implementation details, FGVC performance on iNAT-2021 birds, satellite image to ground level image retrieval performance, and transfer learning performance on CUB-200-2011~\cite{wah2011caltech} and NABirds~\cite{van2015building}.

\subsection{Implementation Details}
We randomly crop the ground-level images to a resolution of 384x384 and satellite images to a resolution of 224x224. We use the ViT-B/32 and ViT-B/16 architecture for the ground-level images and satellite images respectively. For pre-training, we use the AdamW~\cite{loshchilov2017decoupled} optimizer with a weight decay of 0.01. We use a learning rate of 1$e^{-4}$ along with cosine annealing warm restarts~\cite{loshchilov2016sgdr}. We also apply TrivialAugment~\cite{muller2021trivialaugment}.

For linear probing and fine-tuning, 
we use the AdamW optimizer with a weight decay of 1e-4 for linear probing and 0.2 for fine-tuning. The learning rates are set to 0.1 and 5$e^{-5}$ for linear probing and fine-tuning respectively. For fine-tuning, we additionally apply RandAugment~\cite{cubuk2020randaugment}, mixup~\cite{zhang2017mixup}, CutMix~\cite{yun2019cutmix} and LabelSmoothing~\cite{szegedy2016rethinking}.

We use a batch size of 308 across 4 NVIDIA A100 GPUs for all the experiments. Additional details about all our implementations are present in the Appendix.
\subsection{Cross-View iNAT-2021 Birds Experiments}
After pre-training, we do supervised training on the Cross-View iNAT-2021 Birds Dataset to evaluate the representations learned by our proposed models. This is done by reporting linear probing as well as fine-tuning accuracy scores on the standard test set of the Cross-View iNAT-2021 Birds Dataset. For CVE-MAE-Meta and CVM-MAE-Meta models, we add metadata-dependent features to the \textit{[cls]} token before classification with the linear head. Again, we use a dropout of 0.25 for the metadata. We compare the performance of our models with the following baselines: MoCo-V2-Geo\footnote{Please note that we implement our version of cross-view training of MoCo-V2-Geo.}~\cite{ABayush2021geography}, MAE~\cite{he2022masked} and Metaformer-2~\cite{ACdiao2022metaformer}. 

Results illustrated in Table~\ref{table-runtime} show that our single stream CVM-MAE-Meta model beats all other models. The incorporation of satellite images during supervised training has helped the metric-based models beat the embedding-based models. For both training strategies, metadata has improved the testing accuracies by at least 1.57\%. Notice that there is a large gap in accuracies between linear probing and fine-tuning. This suggests that complete fine-grained knowledge about the bird species has not yet been embedded into the models during the pre-training stage. 

\begin{figure*}[!ht]
\begin{center}
   \includegraphics[width=\linewidth]{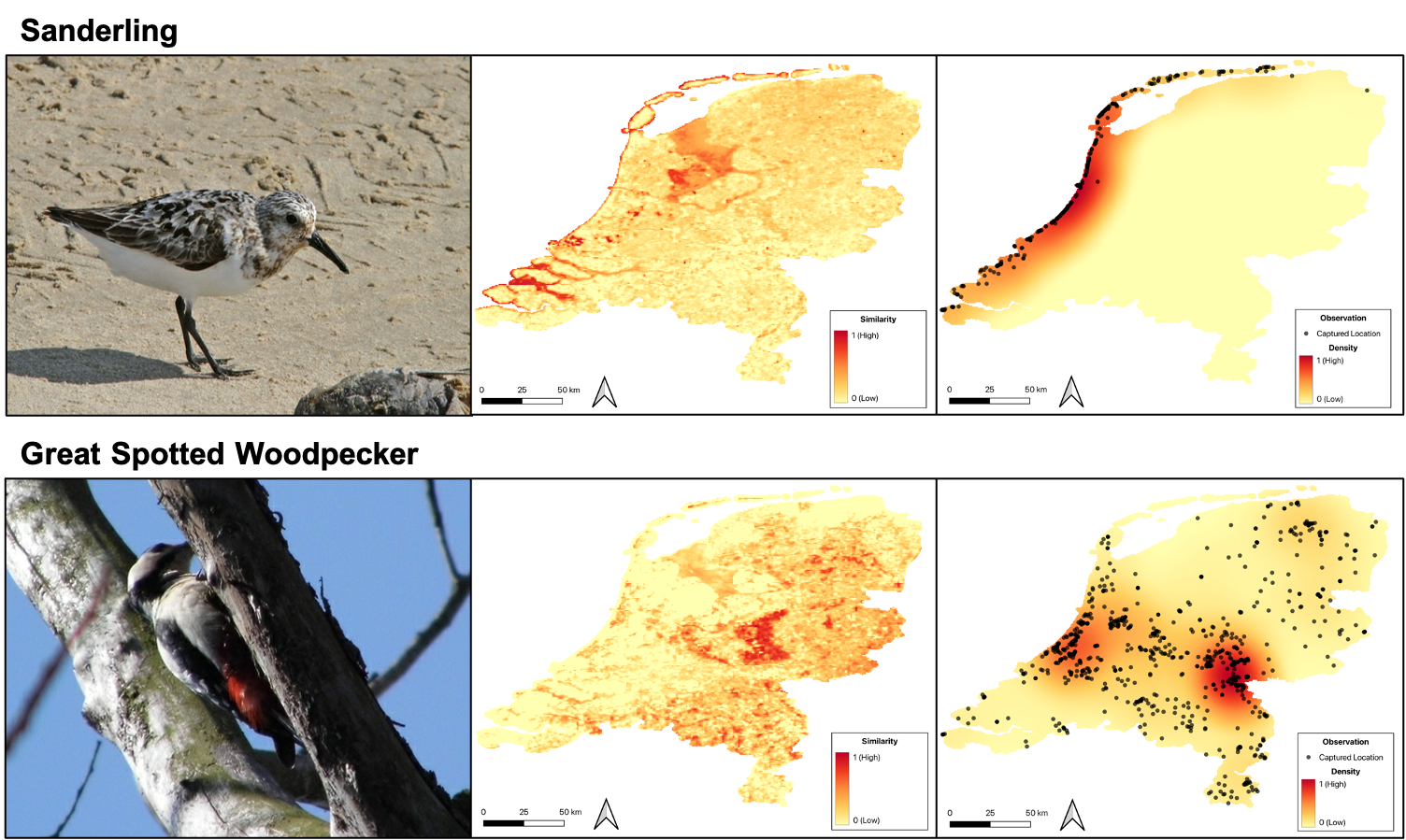}
\end{center}
   \caption{\textbf{Generated Bird Distribution Maps}. Using ground-level bird image to satellite image similarity scores, we generate \textit{expected} bird distribution maps over The Netherlands. We show distribution maps of two different species of birds (left) found in The Netherlands taken from the test set of Cross-View iNAT-2021 Birds Dataset (middle). We also show a heatmap of the presence of those species (right) in the iNAT-2021 Birds Dataset.}
\label{fig:cubrecon}
\end{figure*}

\subsection{Zero-Shot Retrieval}
The zero-shot retrieval experiment contains two subtasks: satellite image to ground-level bird image retrieval and ground-level bird image to satellite image retrieval. We use the former task for computing retrieval metrics while the latter task is for generating species distribution maps. Both retrieval tasks are evaluated using pre-trained models before fine-tuning.

For this first task, we evaluate two different retrieval approaches: 1) Single-stage uni-modal retrieval and 2) Hierarchical cross-modal retrieval. For the single-stage uni-modal retrieval approach, we first compute the [cls] embeddings for all the ground-level bird images and the query satellite image in the test set. Then, we compute the pairwise similarity between the query satellite image embedding and the ground-level image embeddings. Finally, we select the top-k ground-level images with the highest similarity. Additionally, metadata is added for retrieval using our CVE-MAE-Meta model. For calculating the retrieval metrics, we only consider retrieving the correct species rather than the exact image.

The hierarchical cross-modal retrieval approach consists of two stages: 1) selecting candidate ground-level images from similarity computed using the uni-modal model and 2) selecting final ground-level images from the candidates using the matching score computed by the cross-modal model. We propose this approach since embeddings cannot be precomputed for the cross-modal models and are computationally infeasible for retrieval.  

We show the recall scores (R@5 and R@10) of satellite image to ground-level bird image retrieval in Table~\ref{zero-shot}. 
The results indicate that satellite images are able to provide a strong cue indicating that habitat and surroundings are important for the retrieval of bird species. On the other hand, both our dual stream models are able to beat MoCo-V2-Geo model indicating that our models have learned more robust embedding spaces. The hierarchical retrieval approach using CVM-MAE-Meta model also performs reasonably well showcasing the effectiveness of first reducing the search space using uni-modal models and then selecting the final candidates using cross-modal models. However, this approach requires pre-training two separate models and searching steps increasing the computational complexity of the overall setup.

\begin{table}[!b]
\caption{Zero-shot satellite image to ground level bird image retrieval results on the standard test set of Cross-View iNAT-2021 Birds Dataset.}
\label{zero-shot}
\begin{center}
\resizebox{\columnwidth}{!}{%
\begin{tabularx}{0.95\columnwidth}{lccc}
Method & R (@5) &R (@10)&\#FLOPS\\
\midrule
MoCo-V2-Geo~\cite{ABayush2021geography} &5.77&14.28&29.05G\\
\bottomrule
CVE-MAE &14.45&25.62&29.05G\\
CVE-MAE-Meta &13.72&26.97&29.05G\\
CVM-MAE-Meta &\cellcolor{gray!15}\textbf{14.52}&\cellcolor{gray!15}\textbf{28.88}&60.64G
\end{tabularx}}
\end{center}
\end{table}

Figure~\ref{fig:birddist} depicts examples of satellite images retrieved corresponding to query bird images. Clearly, the retrieved images correspond well with the bird's expected habitat. Further, we generate a species distribution map for two distinct species over The Netherlands (Figure~\ref{fig:cubrecon}). We first collected satellite images over a dense grid draped over The Netherlands. We then interpolated and plotted the ground-level image to satellite image similarity scores. We removed all the observations with similarity scores below zero (more details in Appendix Section D).

To study the true performance of our model, we used bird images from the test set of the Cross-View iNAT-2021 Birds Dataset. We conducted a qualitative evaluation of the maps using the observations present in the ground truth. Visually, the maps nicely delineate the presence of birds in the region. Since crowd-sourced datasets are biased towards areas observing high human traffic, the ground truth is not an exhaustive representation of the species distribution. On the other hand, our model is able to provide fine-grained presence of bird species across large geographic regions.    

\subsection{Transfer Learning Experiments}
We evaluate the performance of transfer learning of our model on downstream FGVC Bird datasets. Since satellite images are not available for the majority of the open-sourced datasets, we only study fine-tuning our CVE-MAE-Meta model. We take the best performing CVE-MAE-Meta model on fine-grained classification of iNAT-2021 Birds and then fine-tune it on downstream datasets. We additionally drop metadata during training and inference as the datasets do not include metadata. We consider two of the most popular fine-grained bird classification datasets: CUB-200-2011 and NABirds. The relative scale of these datasets as compared to the iNAT-2021 Birds dataset is presented in Table~\ref{data-stats}.

\begin{table}[!h]
\caption{Various datasets considered in this study.}
\label{data-stats}
\begin{center}
\begin{tabularx}{\columnwidth}{lccc}
Dataset & \#Training & \#Testing & Categories \\
\midrule
iNAT-2021 Birds &414,847&14,860&1486\\
\bottomrule
CUB-200-2011 & 5,994&5,794&200\\
NABirds &23,929&24,633&555\\
\end{tabularx}
\end{center}
\end{table}

\begin{table}[!ht]
\caption{Comparison of accuracy (\%) achieved by CVE-MAE-Meta and SotA approaches on the standard test set of CUB-200-211 and NABirds datasets. We report linear probing (lin) and fine-tuning (ft) accuracy.}
\label{transfer}
\begin{center}
\begin{tabularx}{\columnwidth}{lcccc}
Method & \multicolumn{2}{c}{CUB} & \multicolumn{2}{c}{NABirds}\\
&lin&ft&lin&ft\\
\midrule
MoCo-V2-Geo~\cite{chen2020improved} &81.19&86.91&83.22&89.26\\
MAE~\cite{he2022masked} &80.33&88.46&82.11&89.23\\
MetaFormer-2~\cite{ACdiao2022metaformer}&-&92.40&-&92.70\\
HERBS~\cite{chou2023fine} &-&93.10&-&93.00\\
\bottomrule
CVE-MAE-Meta&\cellcolor{gray!15}\textbf{82.98}&\cellcolor{gray!15}\textbf{93.23}&\cellcolor{gray!15}\textbf{84.21}&\cellcolor{gray!15}\textbf{93.47}\\
\end{tabularx}
\end{center}
\end{table}

Except for HERBS, all the models were first fine-tuned on iNat-2021 birds dataset. The results in Table~\ref{transfer} show that our model outperforms all other transformer-based models including MoCo-V2-Geo, MAE, Metaformer, and HERBS.
The results are consistent for both linear probing and fine-tuning. Our model achieves noticeably high accuracy when linear probing. This indicates that features learned from fine-tuning our pre-trained model on iNAT-2021 Birds are highly robust and transferable.

\section{Discussion and Conclusion}

In this study, we focused on unifying the problem of fine-grained visual classification (FGVC) and mapping of bird species around the world. 
We constructed a cross-view dataset consisting of paired ground-level bird images and satellite images. 
For applications involving ecological mapping and identification of species, satellite images provide spatially correlated topographical information. Therefore, leveraging freely available satellite imagery in SSL enables us to create species maps for any geographic region. Such maps, when augmented with expert knowledge, may be used to refine existing species distribution maps, which otherwise are usually sparse and inaccurate.

We evaluated two architectural frameworks: uni-modal and cross-modal, trained with masked reconstruction and contrastive learning objectives. They differ in the way multi-modal information is fused, which is an essential component for real-time global-scale applications. Uni-modal setup is useful because the modality-specific encoders can be used to pre-compute embeddings. These embeddings can then be used in real-time for a variety of downstream tasks. This becomes essential in large-scale applications. Still, the uni-modal setup is only able to preserve correlated information between the modalities, limiting its performance on recognition tasks. In contrast, the cross-modal setup allows models to learn complementary information coming from a variety of modalities creating rich features useful for complex visual identification tasks. However, the computational complexity of this setup prevents its adoption for large-scale applications. Yet, we conclude that both our training setups are effective at learning general-purpose features that can used for species classification and mapping. Finally, we also presented a two-stage retrieval approach that takes advantage of both the training setups to reduce computation bottleneck. 


Owing to the flexibility of our framework, 
we can easily incorporate other modalities such as text and sound as part of future work. However, one needs to be careful when including several modalities since data collected from un-curated may fail to provide fine-grained information useful for the task of FGVC. Requiring no domain expertise, additional visual modality is easy to collect and proves to be a strong signal for FGVC. Moreover, freely accessible global scale information such as temperature and digital elevation model (DEM) can easily be incorporated into our framework. We hope that this study paves the way for innovative future methods of species distribution modeling using deep learning. 


{\small
\bibliographystyle{ieeetr}
\bibliography{egbib}
}

\end{document}


\title{BirdSAT: Supplementary Material}

\maketitle

\appendix
\begin{figure*}[!h]
\begin{center}
    \includegraphics[width=0.48\linewidth]{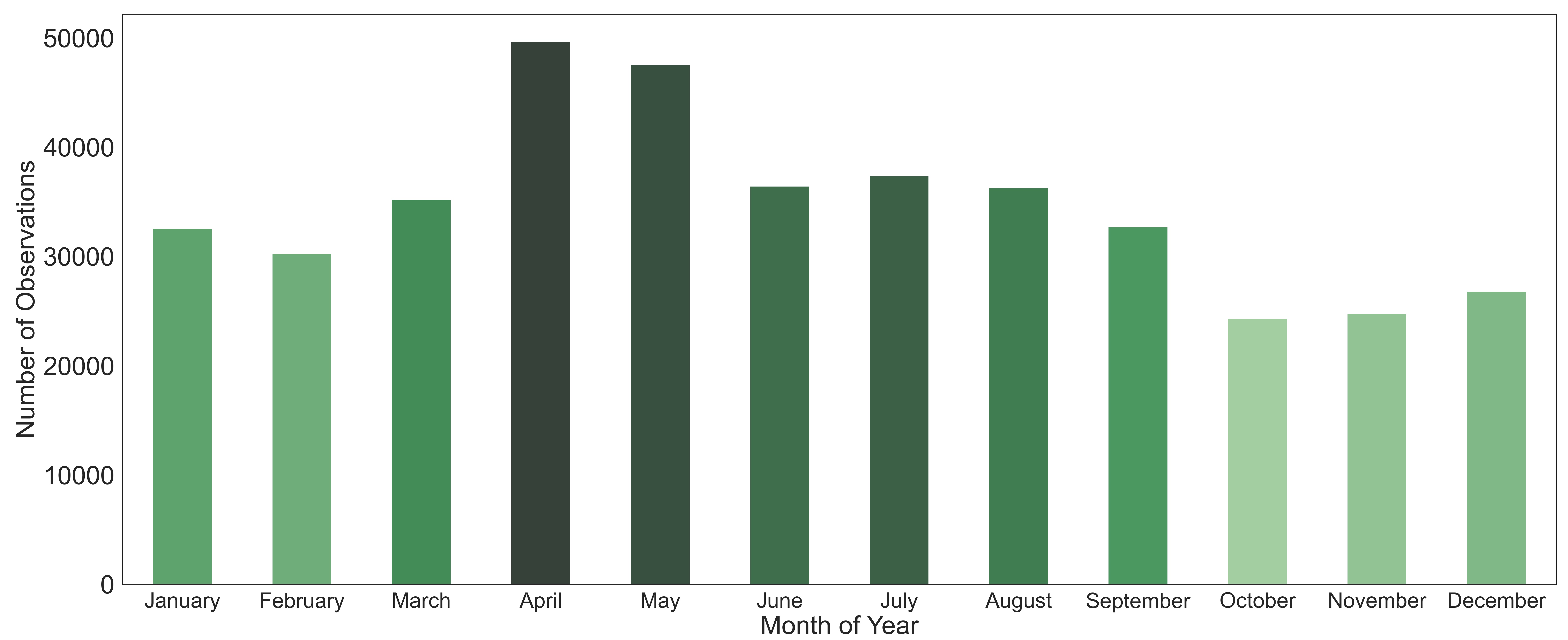}
   \includegraphics[width=0.47\linewidth]{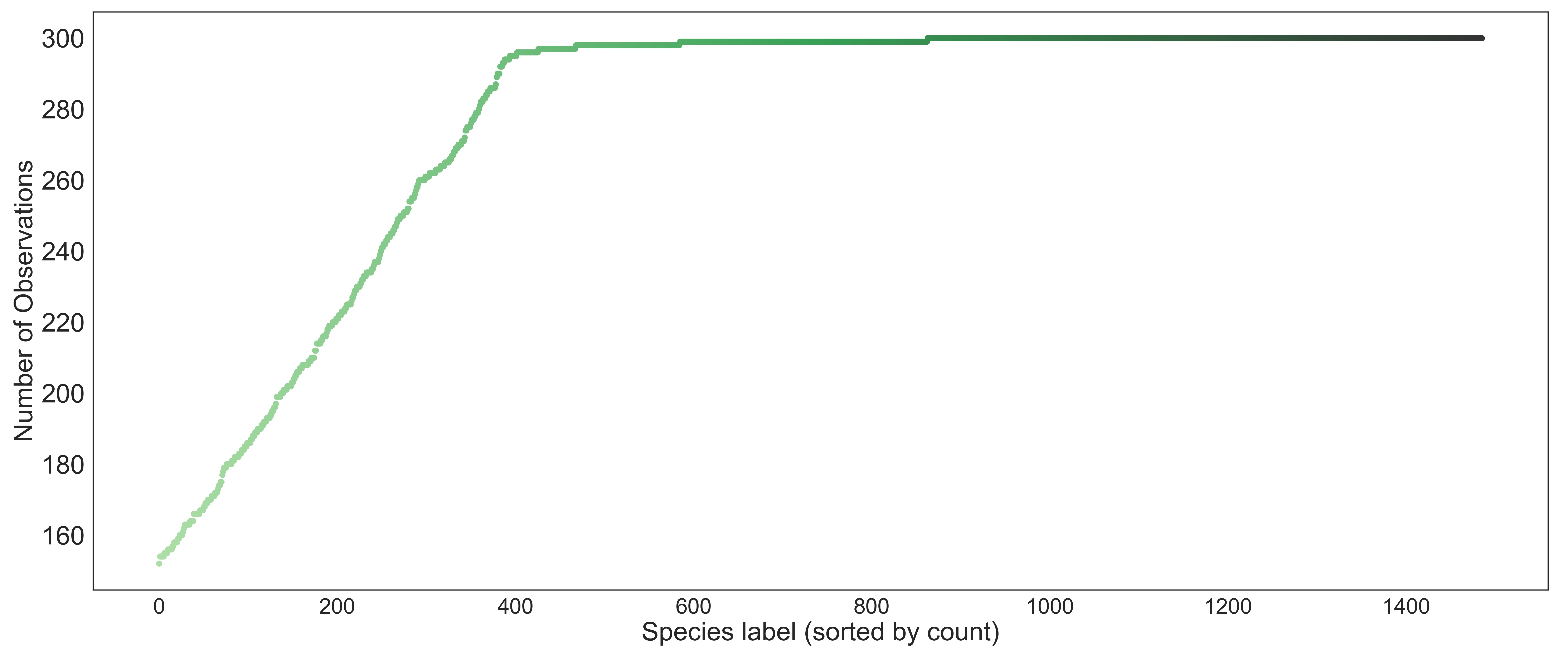}
   
   \caption{\textbf{Distribution of Training Samples}. We show the number of observations per month (left) and number of observations per species (right). The figures indicate that the training set is fairly balanced across the species.}
   \end{center}
\label{fig:countsp}
\end{figure*}
\begin{figure*}[!h]
\begin{center}
    \includegraphics[width=0.48\linewidth]{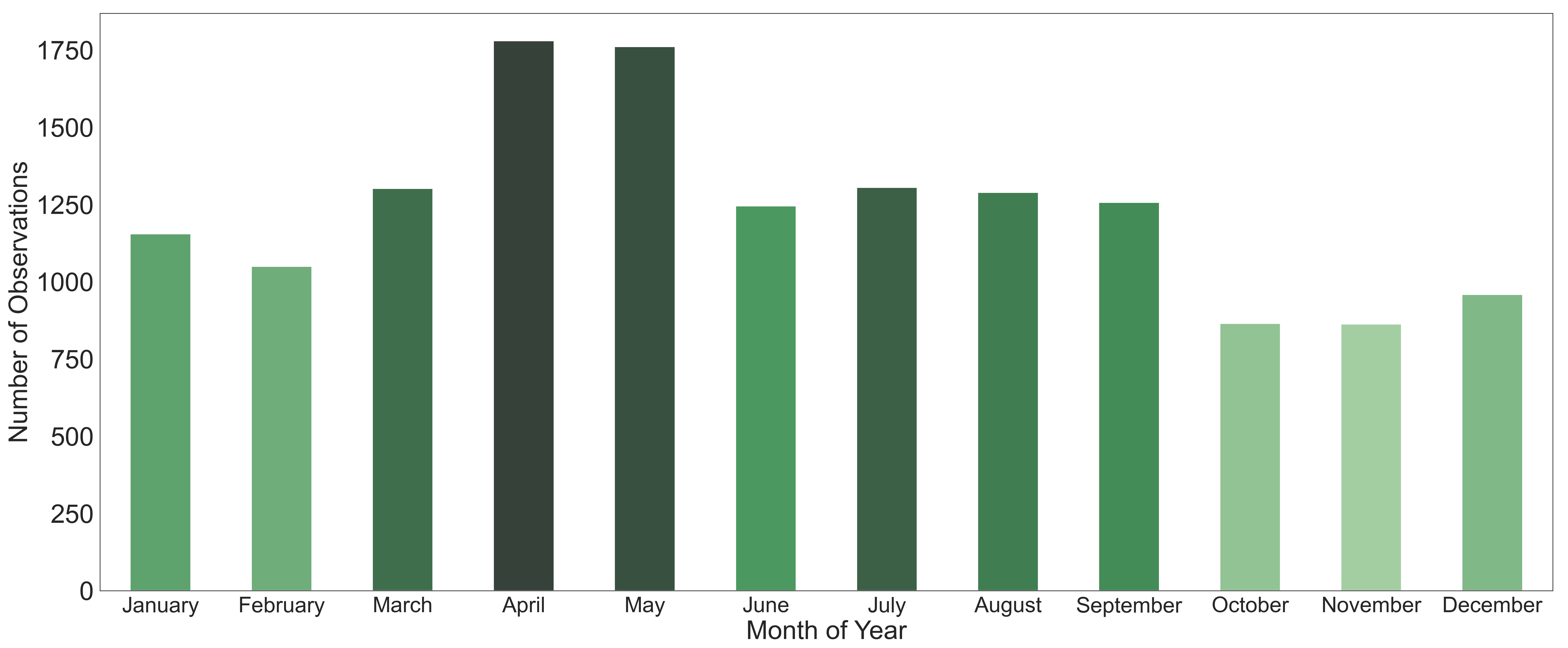}
   \includegraphics[width=0.47\linewidth]{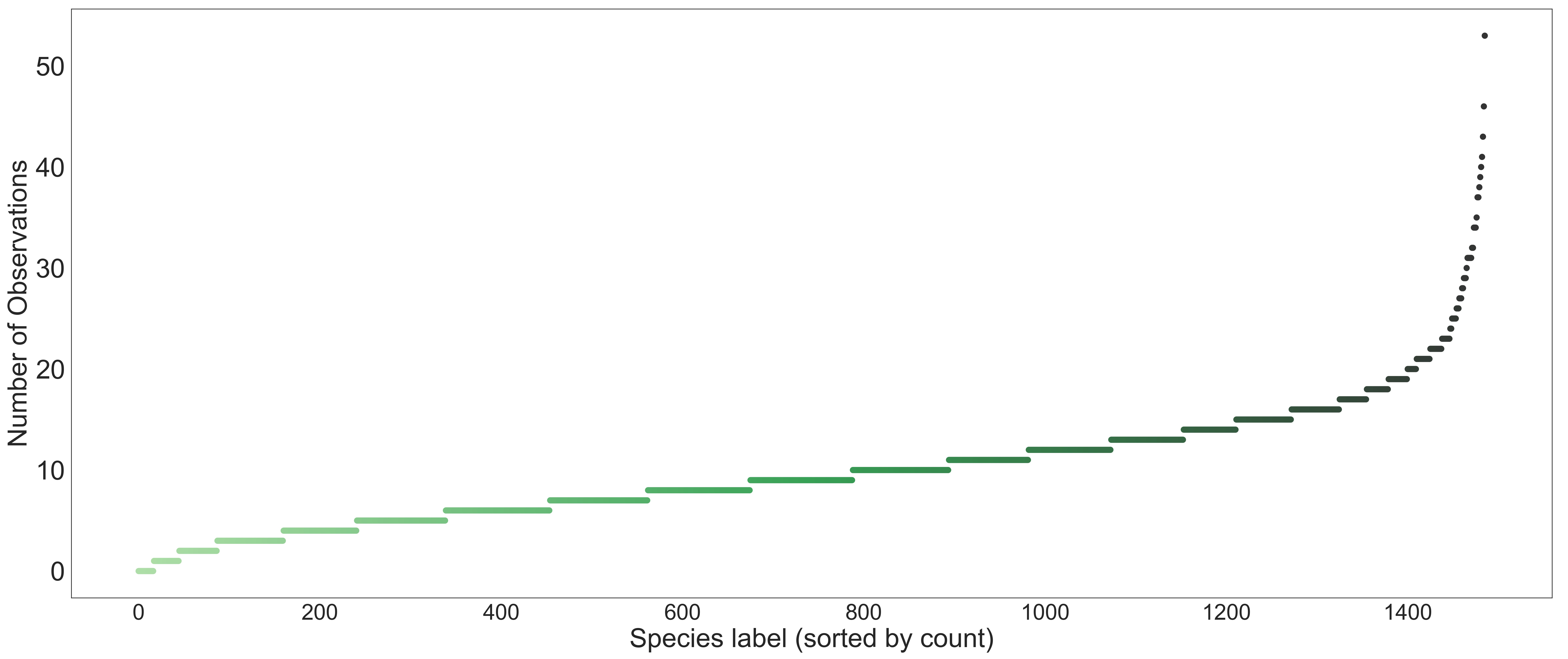}
   
   \caption{\textbf{Distribution of Testing Samples}. We show the number of observations per month (left) and number of observations per species (right). The figures indicate that the distribution of species across the months is fairly balanced while there is a slight imbalance in the number of observations per species.}
   \end{center}
\label{fig:countsp2}
\end{figure*}
\section{Dataset}
\textbf{Data Preparation}. We downloaded the entire iNaturalist 2021 dataset from the iNAT Competition GitHub (\url{https://github.com/visipedia/inat_comp/tree/master/2021}). We sliced the dataset to include only birds using the value \textit{Aves} in category \textit{class} key. This resulted in 414,847 samples in training and 14,680 samples in testing. The maximum and minimum samples per species were 300 and 152 respectively. Then, we applied a \textit{minimal filter} to remove entries with missing geolocations or timestamps. This resulted in dropping only 888 out of 414,847 (0.2\%) observations in training. In testing, we dropped 29 out of 14,860 (0.1\%). 

\textbf{Dataset Details}. The total number of samples in training and testing were 413,959 and 14,831 respectively. The maximum and minimum samples per species after filtering remained the same as before. The distribution of samples per species and per month is shown in Figure~\ref{fig:countsp}1. As is seen, the distribution of samples across the species did not change significantly after filtering. Further, the even distribution of samples across the species comes out-of-the-box from the original dataset. Finally, we collected corresponding satellite imagery for each bird sample using the geolocation present in the dataset. This was done by issuing WMS requests to the Sentinel-2 Cloudless server (\url{https://s2maps.eu/}). 
\section{Training}
\textbf{Training details}. We use the \textit{timm} package for creating all our models and \textit{pytorch\_lightning} (pl) package for training and inference. All the training is done on 4 NVIDIA A100\_SXM4\_40GB GPU's using pl's DistributedDataParellel (ddp) recipe. The experiments are run in parallel across 2 nodes (Intel(R) Xeon(R) Gold 6242 CPU @ 2.80GHz).


\textbf{Metadata}. The iNAT-2021 Birds Dataset includes meta-information in the form of dictionary with keys: \textit{id}, \textit{width}, \textit{height}, \textit{file\_name}, \textit{license}, \textit{rights\_holder}, \textit{date}, \textit{latitude}, \textit{longitude}, \textit{location\_uncertainty}. We extract the \textit{latitude}, \textit{longitude}, and \textit{date} values for each image. For \textit{date}, we extract the \textit{month} and discard all other fields. The three values are mapped as follows: 
\begin{equation}
    lon \rightarrow (sin(\pi*lon/180),  cos(\pi*lon/180))
\end{equation}
\begin{equation}
    lat \rightarrow (sin(\pi*lat/90), cos(\pi*lat/90))
\end{equation}
\begin{equation}
    month \rightarrow (sin(\pi*month/12), cos(\pi*month/12))
\end{equation}
All the values are concatenated and passed to a linear layer which embeds them to a dimension of 768. This embedding is added after extracting features from the encoders. More specifically, we add it to the [cls] token's embedding from the encoders. This final [cls] embedding is used in pre-training and various downstream tasks. If meta-dropout is turned on with probability p, we simply add zeros to the [cls] embedding with probability p during training. 

\textbf{Pre-Training}. We use a masking ratio of 75\% for the masked reconstruction objective. We \textit{do not} use momentum contrast for the contrastive learning objective as~\cite{cmae:huang2022contrastive} only reported a minor improvement in performance. We use color jittering, RandomResizedCrop and RandomHorizontalFlip for the overhead satellite images. For the ground-level images, we only use RandomResizedCrop and TrivialAugment~\cite{muller2021trivialaugment}. The specific details are presented in Table~\ref{param:pre}.

\begin{table}[!t]
\caption{Pre-training hyperparameters and settings.}
\label{param:pre}
\begin{center}
\begin{small}
\begin{tabularx}{\columnwidth}{l|l}
Config & Value\\
\hline
optimizer&AdamW\\
weight decay & 0.01\\
base learning rate & 1e-4\\
batch size & 308\\
optimizer momentum & $\beta_1$=0.9, $\beta_2$=0.95\\
learning rate scheduler & cosine decay\\
input normalization & $\mu$ = [0.485, 0.456, 0.406]\\
&$\sigma$ = [0.229, 0.224, 0.225]\\
masking ratio & 0.75\\
meta dropout & 0.25\\
augmentation\_ground & RandomResizedCrop(384)\\
&  TrivialAugment\\
augmentation\_satellite & RandomResizedCrop(224)\\
& ColorJitter(0.5, 0.5, 0.5)\\
&  RandomHorizontalFlip(p=0.5) 
\end{tabularx}
\end{small}
\end{center}
\end{table}

\begin{table}[!h]
\caption{Linear probing hyperparameters and settings.}
\label{param:lin}
\begin{center}
\begin{small}
\begin{tabularx}{\columnwidth}{l|l}
Config & Value\\
\hline
optimizer&AdamW\\
weight decay & 1e-4\\
base learning rate & 0.1\\
batch size & 308\\
optimizer momentum & $\beta_1$=0.9, $\beta_2$=0.999\\
learning rate scheduler & cosine decay\\
input normalization & $\mu$ = [0.485, 0.456, 0.406]\\
&$\sigma$ = [0.229, 0.224, 0.225]\\
meta dropout & 0.25\\
augmentation\_ground & RandomResizedCrop(384)\\
augmentation\_satellite & RandomResizedCrop(224)
\end{tabularx}
\end{small}
\end{center}
\end{table}

\begin{table}[!h]
\caption{Downstream fine-grained classification hyperparameters and settings.}
\label{param:ft}
\begin{center}
\begin{small}
\begin{tabularx}{\columnwidth}{l|l}
Config & Value\\
\hline
optimizer&AdamW\\
weight decay & 0.1\\
base learning rate & 5e-5\\
batch size & 308\\
optimizer momentum & $\beta_1$=0.9, $\beta_2$=0.999\\
learning rate scheduler & cosine decay\\
input normalization & $\mu$ = [0.485, 0.456, 0.406]\\
&$\sigma$ = [0.229, 0.224, 0.225]\\
meta dropout & 0.25\\
augmentation\_ground & RandomResizedCrop(384)\\
& RandAugment(10, 12)\\
& CutMix = 1.0\\
& mixup = 0.8\\
& LabelSmoothing = 0.1\\
augmentation\_satellite & RandomResizedCrop(224)\\
& ColorJitter(0.5, 0.5, 0.5)\\
&  RandomHorizontalFlip(p=0.5)
\end{tabularx}
\end{small}
\end{center}
\end{table}

\textbf{Linear Probing}. Following~\cite{he2022masked}, we only use RandomResizedCrop during linear probing. All embeddings are normalized before passing to linear layer for classification. Other details are illustrated in Table~\ref{param:lin}. As reported by~\cite{he2022masked}, linear probing accuracy is uncorrleated from fine-tuning accuracy. This explains the fact that there is a large gap in the metrics on iNAT-2021 Birds Dataset. They also concluded that contrastive-based models were better than MAE at linear probing. The combination of contrastive and masked reconstruction objectives allows our model to learn robust features for a variety of downstream tasks and beat purely contrastively trained models.

\begin{table*}[!h]
\caption{Comparison of F1 Score, Precision and Recall achieved by our proposed models and SotA approaches on the standard test set of iNAT-2021 Birds dataset. We report this for fine-tuned models.}
\label{table-runtime}
\begin{center}
\begin{tabularx}{0.82\linewidth}{lcccccc}
Method &Location & Date &Pre-training& F1 Score& Precision & Recall \\
\midrule
MoCo-V2-Geo&\Checkmark&\XSolidBrush&InfoNCE+Geo-Clf.&0.507&0.511&0.503\\
MAE &\XSolidBrush&\XSolidBrush&Recons. Loss&0.488&0.482&0.495\\
MetaFormer-2 &\Checkmark&\Checkmark&ImageNet Clf.&0.510&0.534&0.488 \\
\bottomrule
CVE-MAE & \XSolidBrush& \XSolidBrush&InfoNCE+Recons. Loss& 0.520&0.519 &0.522\\
CVE-MAE-Meta & \Checkmark&\Checkmark &InfoNCE+Recons. Loss&0.527&0.523&0.531\\
CVM-MAE &\XSolidBrush&\XSolidBrush&Matching+Recons. Loss&0.545&0.552&0.539\\
CVM-MAE-Meta &\Checkmark&\Checkmark&Matching+Recons. Loss&\cellcolor{gray!15}\textbf{0.553}&\cellcolor{gray!15}\textbf{0.561}&\cellcolor{gray!15}\textbf{0.546}\\
\end{tabularx}
\end{center}
\end{table*}

\textbf{Fine-Tuning}. For ViT's, fine-tuning (in general) requires severe data augmentations and higher weight decay. As a result, we use RandomResizedCrop, RandAugment, CutMix, mixup and LabelSmoothing. Other details are preseneted in Table~\ref{param:ft}. We also report additional metrics of our fine-tuned models in Table~\ref{table-runtime}.

\begin{figure}[!ht]
\begin{center}
   \includegraphics[width=\columnwidth]{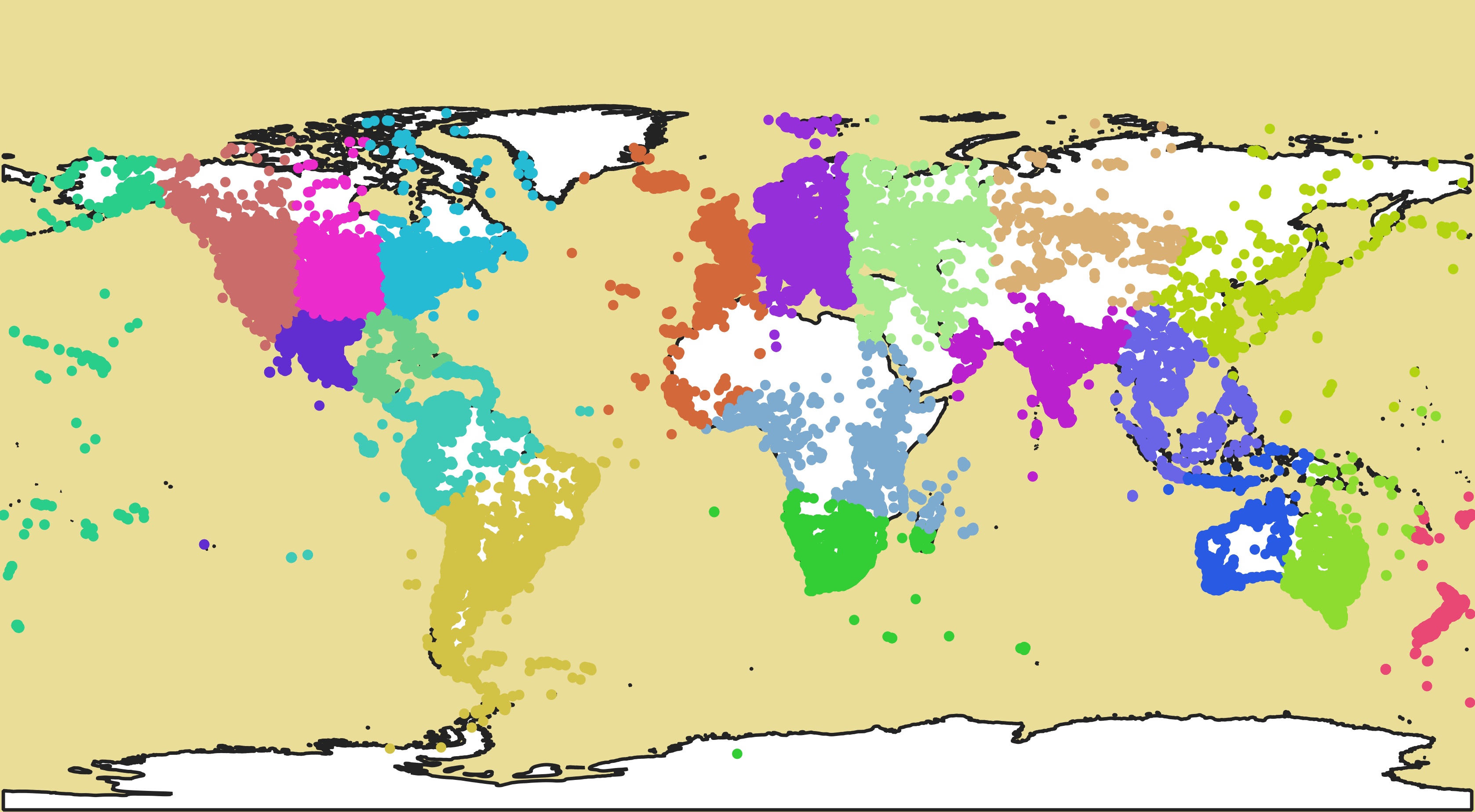}

   \vspace{5mm}
   
   \includegraphics[width=\columnwidth]{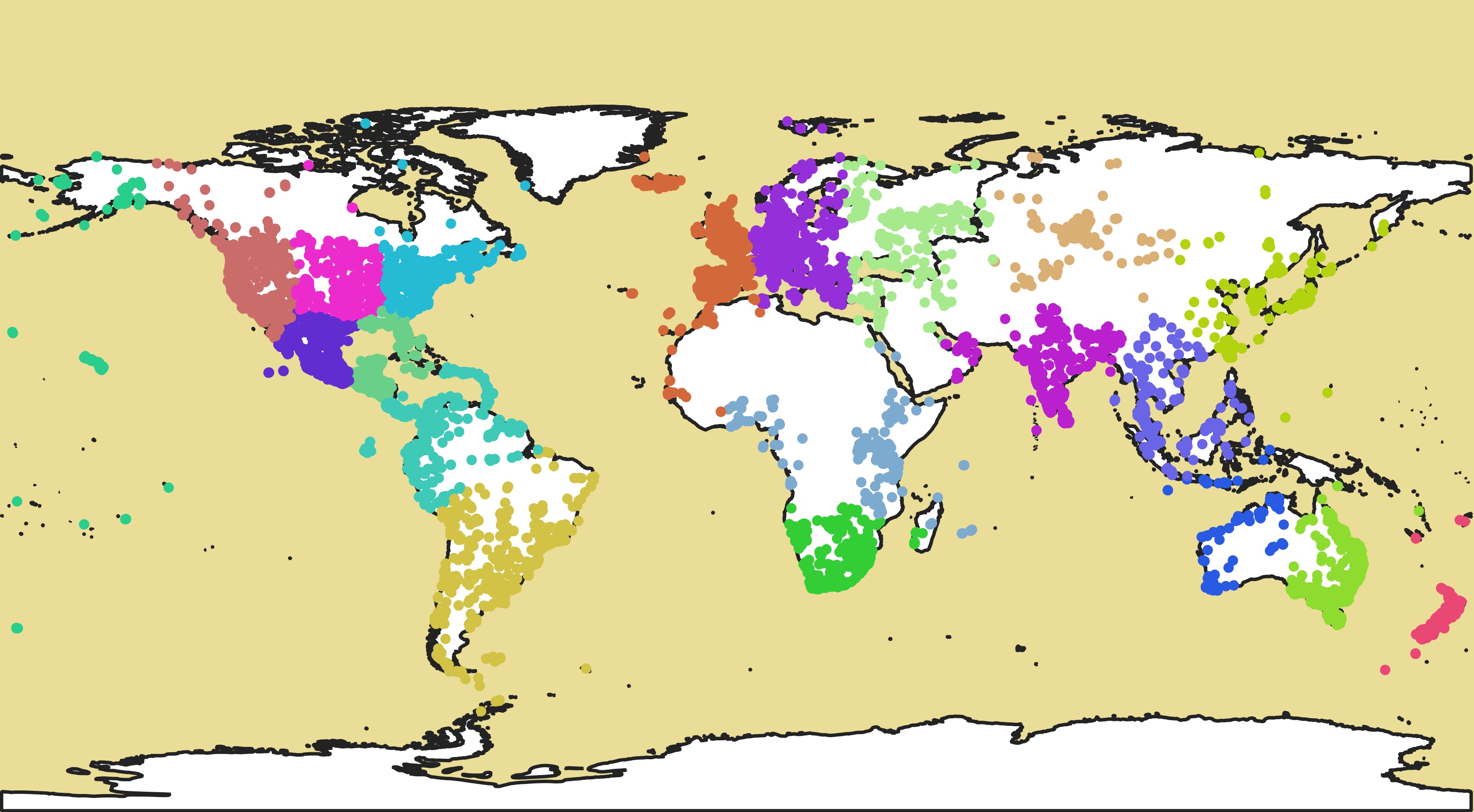}
   
   \caption{\textbf{Geo-Clusters}. Using KMeans Clustering, we cluster the geographic coordinates into 20 classes based on latitude and longitude values. These classes are then used for training on the geo-classification objective for the MoCo-V2-Geo method. Here, we show the training (top) and testing (bottom) geo-locations of images along with color representing their cluster label. }
   \end{center}
\label{fig:clus}
\end{figure}

\textbf{MoCo-V2-Geo}. To make fair comparisons, we implemented a cross-view training routine for the MoCo-V2-Geo. Instead of utilizing temporal positives or data augmentation techniques to create positive and negative pairs, we use the corresponding satellite images. We cluster the geographic coordinates present in the meta-information into 20 classes using the KMeans clustering algorithm (Figure~\ref{fig:clus}). These labels are then used for computing the geo-classification loss using the [cls] embeddings obtained from the ground-level image encoder. We use a queue of size 10000 and the same data augmentations for the ground-level and satellite images as reported in Table~\ref{param:pre}.

\section{Ablation Study}
We conduct ablation on the meta-dropout rate which is a key component of our architecture. It tunes the dependence of models on metadata. Rest of the components are pre-trained SotA architectures which have been extensively studied in previous literatures. As is seen in Table~\ref{meta-abla}, models severely overfit on metadata, when meta-dropout is turned off. Also, the performance of the models' decrease as meta-dropout rate increases more than 25\%. 
\begin{table}[!h]
\caption{Impact of meta-dropout rate on the classification performance of models on Cross-View iNAT-2021 dataset.}
\label{meta-abla}
\begin{center}
\resizebox{\columnwidth}{!}{
\begin{tabularx}{1.08\columnwidth}{lccccc}
Model &  \multicolumn{5}{c}{Meta-Dropout Rate}\\
&0.00&0.25&0.50&0.75&1.00\\
\midrule
CVE-MAE-Meta & 82.23&\textbf{86.23}&85.02&84.79&83.78\\
CVM-MAE-Meta &83.55&\textbf{87.46}&86.22&85.97&85.89\\
\end{tabularx}}
\end{center}
\end{table}
\section{Species Distribution Mapping}

Species distribution maps are constructed by first collecting satellite images over a dense rectangular grid draped on the area of interest. Then, similarity scores are computed between a query bird image and the satellite images. More specifically, we precompute the embeddings for all the image pairs on GPU and then compute their similarity on CPU. The grid of scores is then interpolated at a desired spatial resolution. We use the IDW interpolation and a spatial resolution of $0.01^o$ for The Netherlands. Further, we clamp negative similarity scores to zero before visualizing. A similar procedure can be followed if one wants to use land cover maps, digital elevation models (DEM), and so on for creating species distribution maps. In the future, one could also incorporate text descriptions as query for generating these maps.

\section{Reconstruction Results}

\begin{figure*}
\begin{center}
   \includegraphics[width=\linewidth]{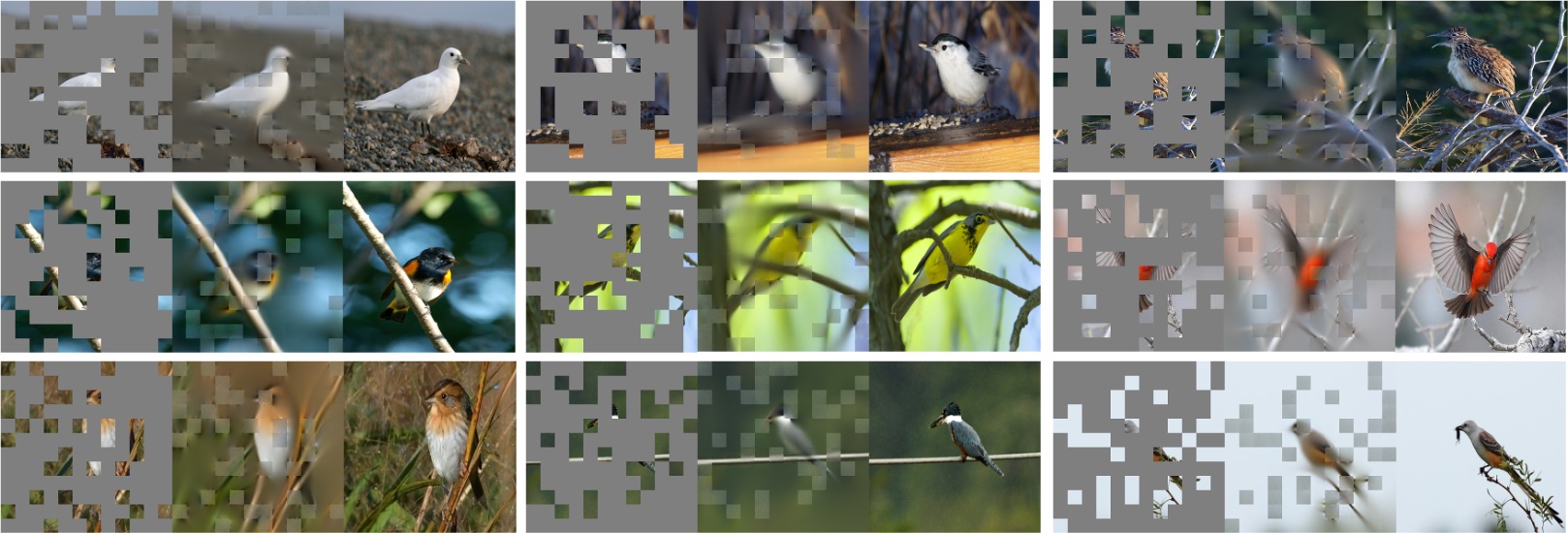}
\end{center}
   \caption{\textbf{Zero-shot reconstruction on CUB-200-2011}. Using pretrained CVE-MAE-Meta, we show zero-shot reconsturction results on randomly selected images from testing set of CUB-200-2011. We show masked (left), predicted (middle) and ground truth (right) images.}
\label{fig:cubrecon}
\end{figure*}

\begin{figure*}[!ht]
\begin{center}
   \includegraphics[width=\linewidth]{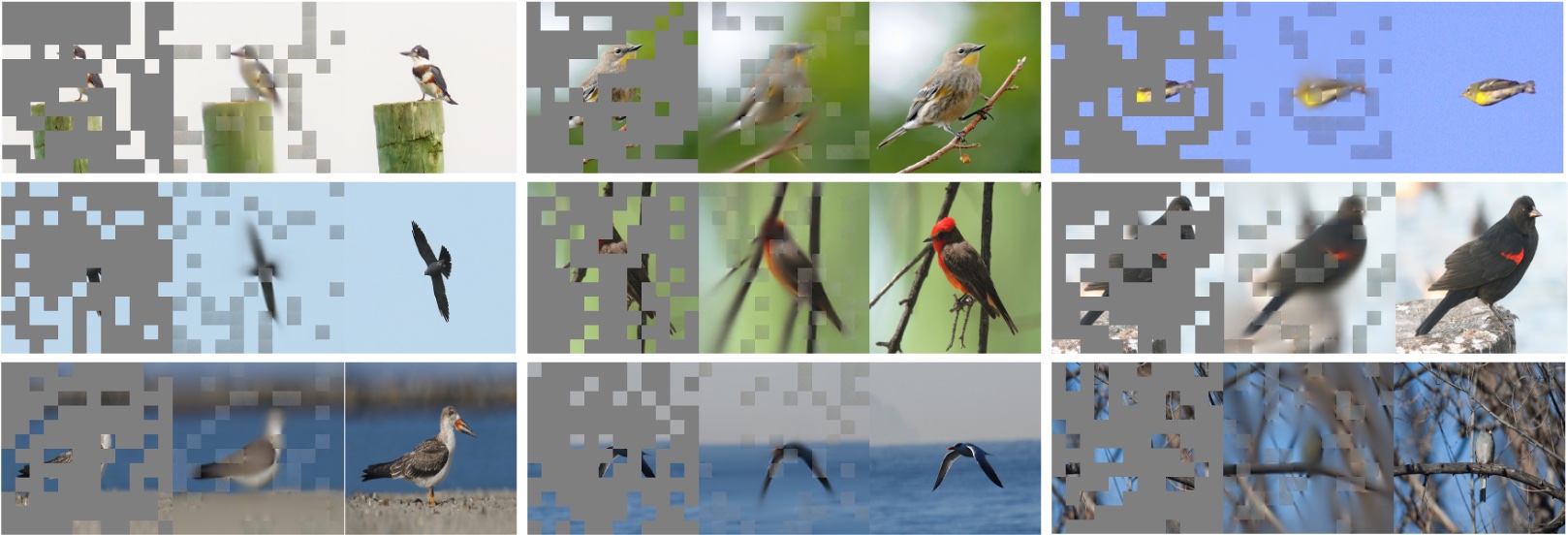}
\end{center}
   \caption{\textbf{Zero-shot reconstruction on NABirds}. Results on randomly selected images from testing set of NABirds. We illustrate masked (left), predicted (middle) and ground truth (right) images.}
\label{fig:narecon}
\end{figure*}

The results of our reconstructions are not fully imperative for species classification and mapping, since our models are trained with a contrastive objective. In the wild, animals come in different poses and sizes, making it challenging to reconstruct them perfectly. However, our models have effectively learned the structures of different bird species. The large scale pre-training along-with satellite imagery and metadata such as month of year and location helps our model learn robust fine-grained features. The zero-shot reconstruction results on CUB-200-2011 and NABirds (Figure \ref{fig:cubrecon} and Figure \ref{fig:narecon}) confirm that the features learned by our models are highly transferable.

\begin{figure*}
\begin{center}
   \includegraphics[width=\linewidth]{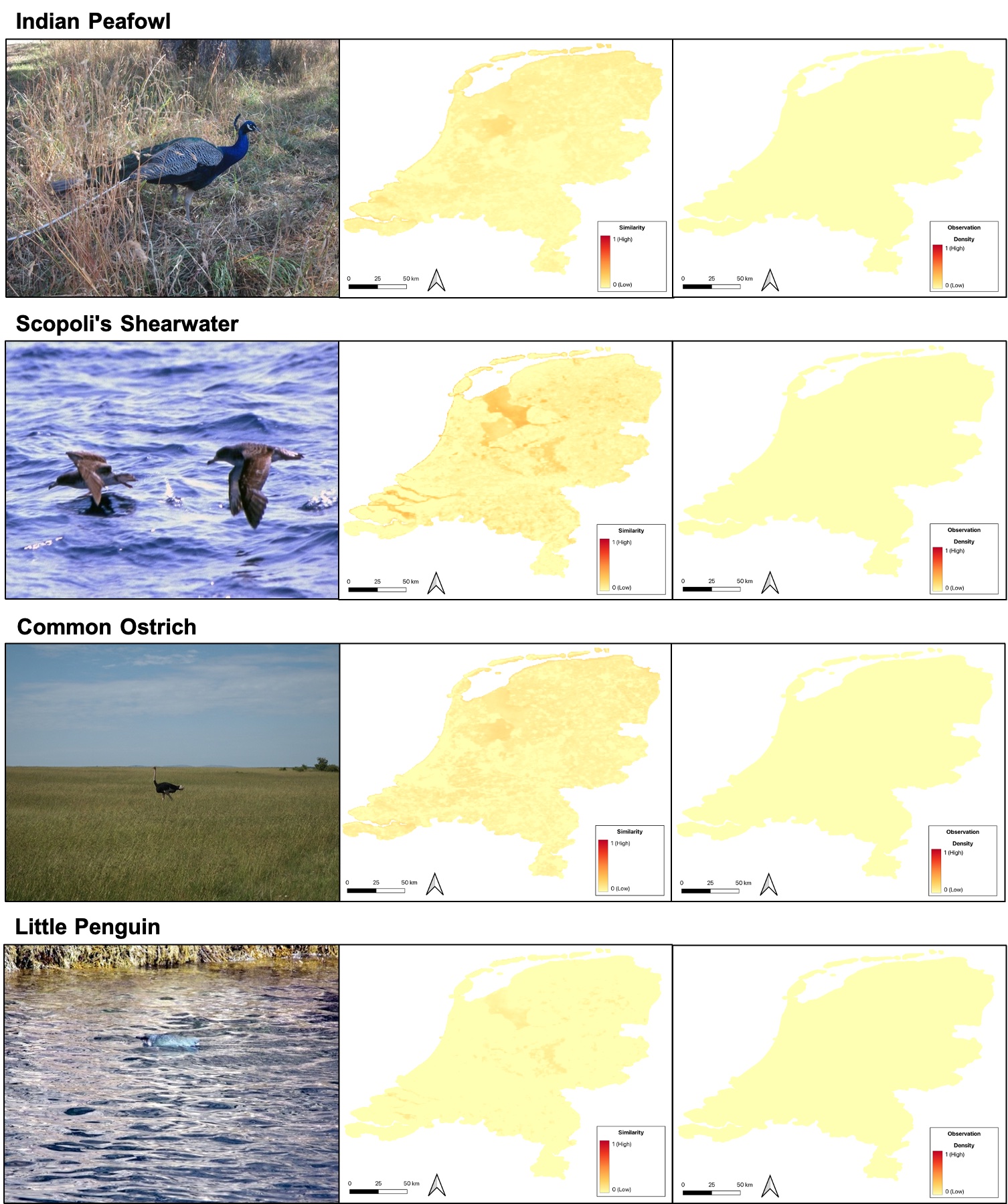}
   
   \caption{\textbf{Bird Maps for Negative Queries}. We select four bird species which are not typically found in The Netherlands. We show ground-level to satellite image similarity scores for those bird species over The Netherlands. Clearly, the maps show little to no activations.}
   \end{center}
\label{fig:negsim}
\end{figure*}
\begin{figure*}
\begin{center}
   \includegraphics[width=\linewidth]{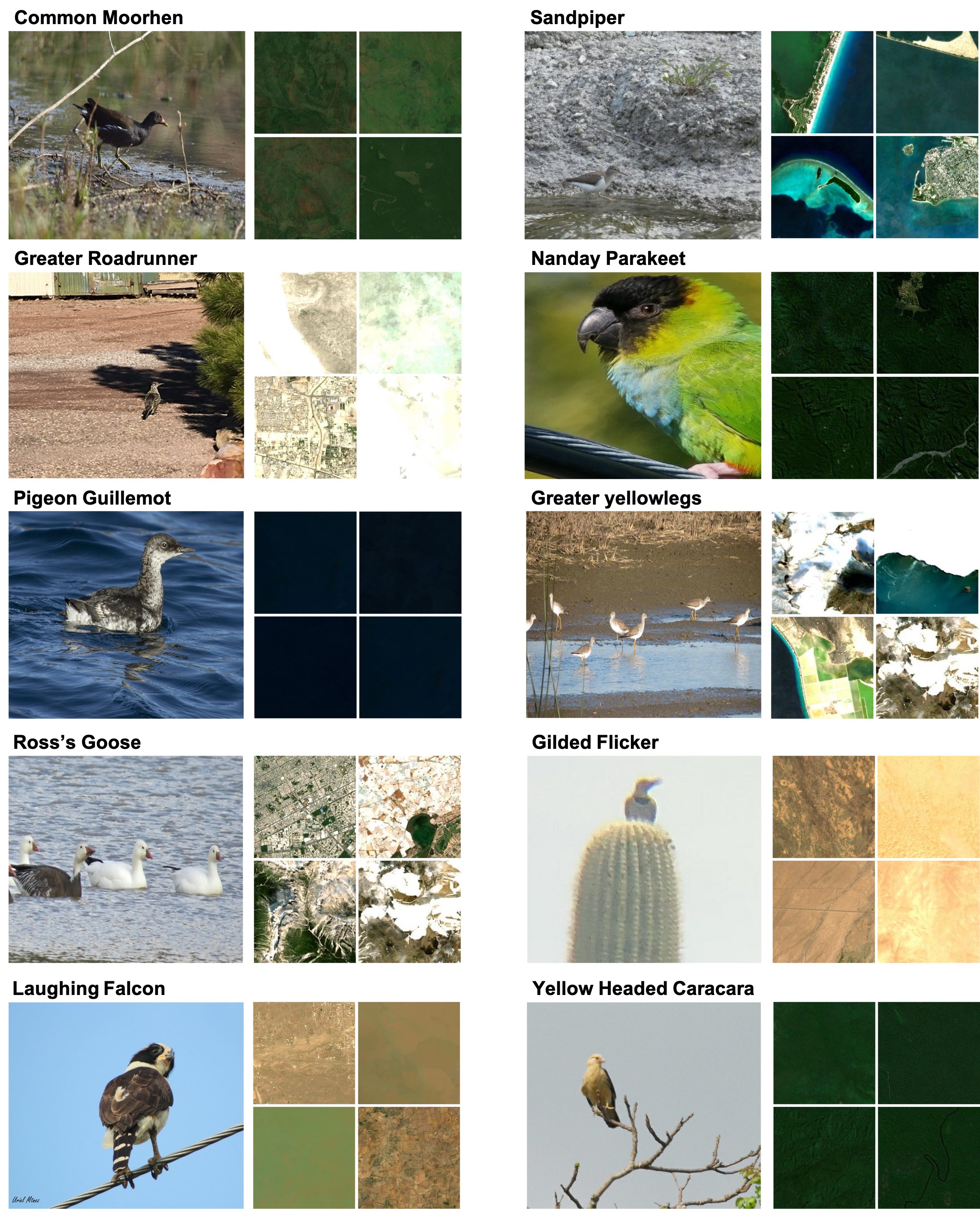}
   \caption{\textbf{Bird to Satellite Image Retrieval}. We show additional examples of uni-modal bird to satellite image retrieval. Clearly, our modal is able to associate bird species with their expected habitat and location.}
   \end{center}
\label{fig:retsupp}
\end{figure*}
\begin{figure*}
\begin{center}
   \includegraphics[width=\linewidth]{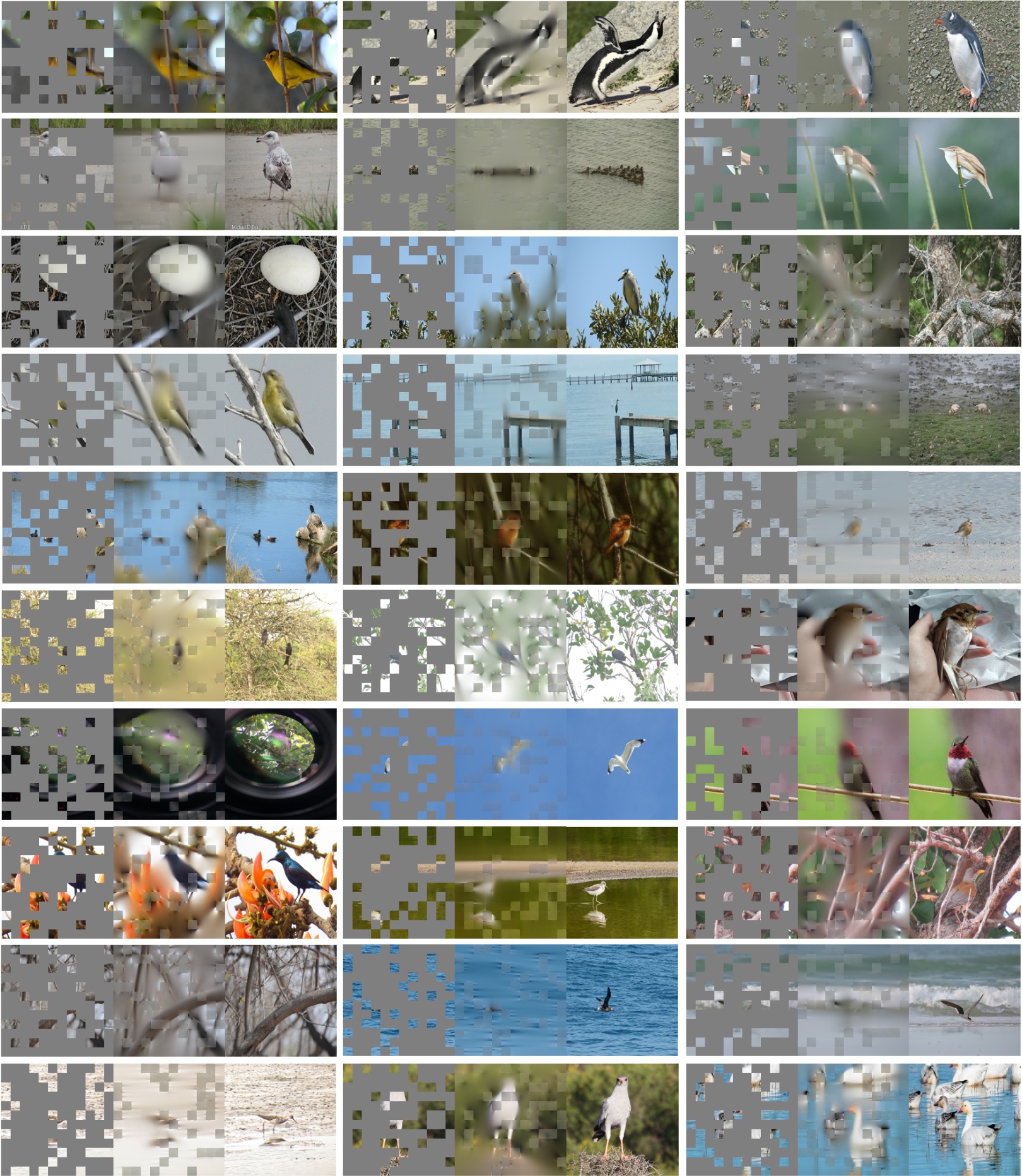}
   
   \caption{\textbf{Reconstruction Results}. Using pre-trained cross-view metric MAE model, we show reconstruction results on randomly selected images from the standard test set of Cross-View iNAT-2021 Birds Daataset. We illustrate masked (left), predicted (middle) and ground truth (right) images. The masking ratio is fixed at 75\% during the inference.}
   \end{center}
\label{fig:recons}
\end{figure*}


{\small
\bibliographystyle{ieeetr}
\bibliography{egbib}
}